\pdfoutput=1

\documentclass[11pt]{article}

\usepackage{acl}

\usepackage{times}
\usepackage{latexsym}

\usepackage[T1]{fontenc}

\usepackage[utf8]{inputenc}

\usepackage{microtype}

\usepackage{inconsolata}

\usepackage{enumitem}
\usepackage{times}
\usepackage{latexsym}
\usepackage{graphicx}
\usepackage{svg}
\usepackage{tabularx}
\usepackage{hyperref}
\usepackage{amsmath, amssymb, bm}
\usepackage{booktabs}

\title{Mitigating Clickbait: An Approach to Spoiler Generation Using Multitask Learning}

\author{Sayantan Pal , Souvik Das , Rohini K. Srihari \\
        State University of New York at Buffalo \\
        Department of Computer Science and Engineering \\
        \texttt{\href{mailto:spal5@buffalo.edu}{spal5}},
        \texttt{\href{mailto:souvikda@buffalo.edu}{souvikda}},
        \texttt{\href{mailto:rohini@buffalo.edu}{rohini}}@buffalo.edu}

\begin{document}
\maketitle
\begin{abstract}
This study introduces 'clickbait spoiling', a novel technique designed to detect, categorize, and generate spoilers as succinct text responses, countering the curiosity induced by clickbait content. By leveraging a multi-task learning framework, our model's generalization capabilities are significantly enhanced, effectively addressing the pervasive issue of clickbait. The crux of our research lies in generating appropriate spoilers, be it a phrase, an extended passage, or multiple, depending on the spoiler type required. Our methodology integrates two crucial techniques: a refined spoiler categorization method and a modified version of the Question Answering (QA) mechanism, incorporated within a multi-task learning paradigm for optimized spoiler extraction from context. Notably, we have included fine-tuning methods for models capable of handling longer sequences to accommodate the generation of extended spoilers. This research highlights the potential of sophisticated text processing techniques in tackling the omnipresent issue of clickbait, promising an enhanced user experience in the digital realm.
\end{abstract}


\section{Introduction}

In today's fast-paced digital world, 'clickbait'—enticing but often misleading headlines designed to generate clicks have become a common trend.\cite{Potthast2016ClickbaitD, sterz-etal-2023-ml}. While these headlines draw users in, they often fail to deliver their promises, leading to user dissatisfaction \cite{10.1145/3411764.3445753}. Consequently, 'clickbait spoiling'—providing succinct, truthful summaries (or 'spoilers') of clickbait content—has emerged as a powerful strategy to counter this trend \cite{hagen-etal-2022-clickbait}.

\begin{figure}[h]
\centering
\includegraphics[width=0.30\textwidth]{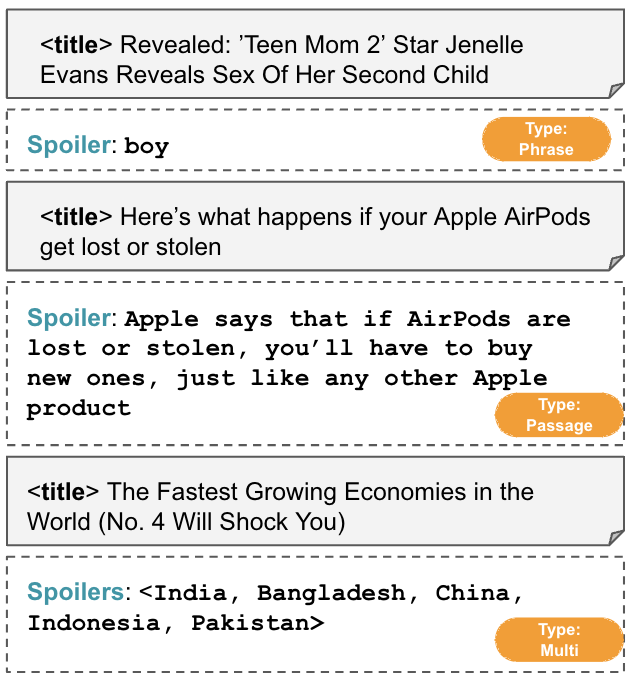}
\caption{Examples of different categories of clickbait spoilers in Webis \cite{hagen-etal-2022-clickbait} dataset.}
\label{fig:cb_ex}
\end{figure}

However, detecting, categorizing, and generating these spoilers involves sophisticated Natural Language Processing (NLP) techniques \cite{Kurenkov2022SavedYA}. Furthermore, the type of required spoiler can vary from a short phrase to an extended passage or multiple snippets from the document, adding another layer of complexity to the problem \cite{48643}. Examples illustrating each of these types of spoilers are presented in Figure \ref{fig:cb_ex}. \citet{hagen-etal-2022-clickbait} has proposed a two-step methodology to tackle clickbait. Although, \citeauthor{hagen-etal-2022-clickbait} approach is a crucial contribution, it has enough room for improvement. Our research aims to extend their work by exploring ways to identify passage and multi-spoilers and enhance the user experience by providing accurate spoilers to clickbait content, mitigating the negative impacts of such misleading headlines \cite{10.5555/3192424.3192427}.

\textbf{Contribution}: In this study, we introduce a simple and novel methodology to solve the clickbait spoiling task using multi-task learning(MTL) framework,a method that enables the simultaneous learning of multiple related tasks to improve performance across them as outlined by Liu et al. \cite{liu-etal-2019-multi}. This allows a more comprehensive representation of the problem space, potentially enhancing its ability to generate accurate and diverse spoilers from different aspects of the same content. Specifically, we show that using \textbf{Spoiler generation in MTL setting can increase the overall generation quality}. Furthermore, we show \textbf{finetuning LongT5 model \cite{48643} has displayed superior performance, achieving 60\% higher BLEU-4 over our multi-task approach.}

\section{Background: Clickbait Spoiling}
Clickbait spoiling has garnered significant attention in the research community, primarily driven by the increasing prevalence of clickbait headlines\cite{hurst2016clickbait} in the digital media landscape \cite{hovy-etal-2013-learning}. The misleading headlines \cite{Wei2017LearningTI} often lead to user dissatisfaction due to the discrepancy between the promised and delivered content \cite{potthast2016clickbait}. It has prompted the emergence of clickbait spoiling, a strategy aimed at providing users with honest, succinct summary generation techniques\cite{Pal2021SummaryGU} to generate spoilers of the clickbait content, thus fostering a more satisfying user experience \cite{hagen-etal-2022-clickbait}.

\subsection{Existing Approaches}
\citet{10.1007/978-3-319-30671-1_72} and \citet{7877426} have explored clickbait detection using Classical Machine Learning (ML) and Deep Learning (DL) approaches.\citet{hagen-etal-2022-clickbait} delves into more challenging clickbait spoiling in a unique two-step approach, which involves classifying spoiler types by finetuning BERT\cite{devlin-etal-2019-bert} based models and treating the spoiler generation task as a question-answering or passage retrieval\cite{karpukhin-etal-2020-dense} challenge. Lastly, it comprehensively evaluates leading techniques for spoiler-type classification and passage retrieval. The paper suggests that spoiler-type classification can be beneficial but not crucial and indicates an unaddressed area of research for generating multi-part spoilers.
\subsection{Data}
\hypertarget{item:a}{}
We used the Webis Clickbait\cite{hagen-etal-2022-clickbait} Spoiling Corpus 2022\footnote{Data: https://webis.de/data.html?q=clickbait} to conduct our experiments, it is a collection of 5,000 spoiled clickbait posts gathered from social media platforms like Facebook, Reddit, and Twitter. This corpus aims to assist in clickbait spoiling\cite{Kurenkov2022SavedYA}, a process that involves creating a brief piece of text that satiates the curiosity generated by a clickbait post. The corpus is divided into 3,200 training samples, 800 validation samples, and 1,000 testing samples, which consist of phrases, passages, and multi-spoilers.

\section{Our Approach}

\begin{figure*}[ht]
    \centering
    \includegraphics[width=0.75\textwidth]{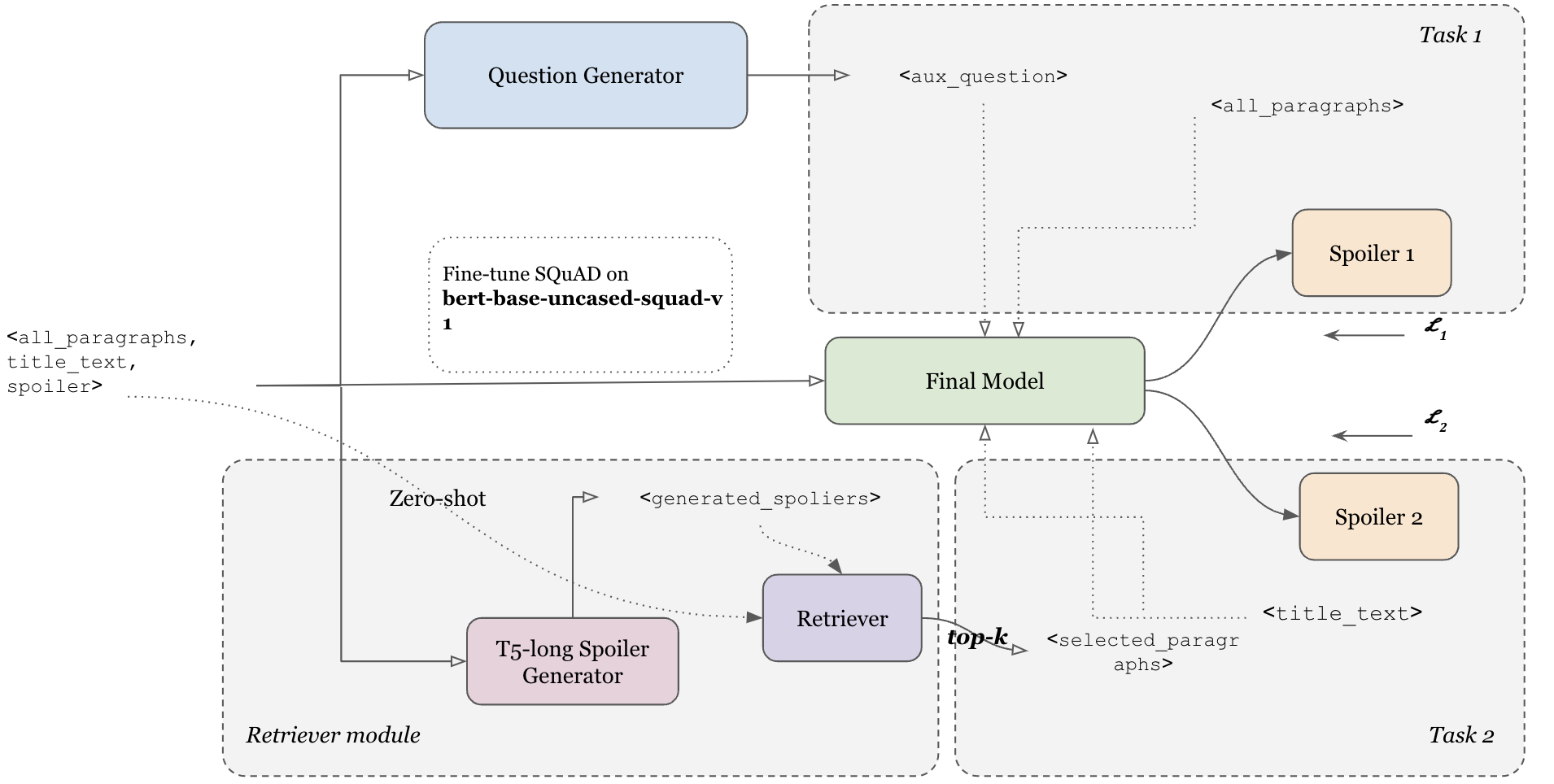}
    \caption{Overview of the Multitask Learning Learning Spoiler Generation Model}
    \label{fig:spoiler-gen-mtl}
\end{figure*}

We address the challenge of mitigating clickbait by a two-step process. The first task is to classify the spoilers into phrases, passages, or multi, followed by the second task of generating contextually accurate and informative spoilers through a multi-task learning framework.

\subsection{Clickbait Spoiler Classification}
\label{classification}

The primary goal of our spoiler classification approach is to analyze a document and accurately categorize spoilers based on their length and context, identifying them as short phrases, longer passages, or spanning multiple sections. In our spoiler classification approach, we leverage both the paragraphs (context) and the title text (question), which are concatenated and formatted as: \texttt{<[title\_text] [SEP] [context]>} as the input, with the golden spoiler class labels (0, 1, 2) indicating phrase, passage, and multi, respectively. To process this input, we employ a 512-token sequence, applying necessary padding and truncation to maintain a consistent input size. A BERT-based encoder\cite{devlin-etal-2019-bert} transforms this input into dense vector representations, mainly focusing on the hidden representation of the \texttt{[CLS]} token, encapsulating global information from the input. This representation is then passed through a linear layer followed by a softmax activation function to perform the classification. Cross-entropy loss given in equation \ref{eq:cross_entropy_loss} optimizes the model parameters, ensuring accurate and reliable spoiler detection.

\begin{equation}
L(y, \hat{y})=-\sum_{i=1}^C y_i \log \left(\hat{y}_i\right)
\label{eq:cross_entropy_loss}
\end{equation}

\subsection{Clickbait Spoiler Generation}

Our objective is to perform Question Answering to generate the spoilers from the documents. In our multi-task learning framework for spoiler generation, we process paragraphs \( \bm{P} \in \mathbb{R}^{d \times P} \), title text \( \bm{Q} \in \mathbb{R}^{d \times Q} \), and golden spoilers \( \bm{A} \in \mathbb{R}^{d \times A} \) as inputs, aiming to generate spoilers by Question Answering. Here, \( d \) represents the dimensionality of the embeddings, and \( P \), \( Q \), and \( A \) represent the lengths of these sequences. Our multi-task framework consists of the following tasks:

\textbf{Task 1: Spoiler Generation with Auxiliary Question}
In this task, we guide the spoiler generation using auxiliary questions generated using a question generator\footnote{\href{https://huggingface.co/valhalla/t5-base-qg-hl}{valhalla/t5-base-qg-hl}}. The task can be formulated as:
\begin{equation}
\bm{S}_1 = f_{\text{QA}}(\bm{P}, \bm{Q}_{\text{aux}}, \bm{A})
\end{equation}
where \( \bm{S}_1 \) is the generated spoiler, \( \bm{Q}_{\text{aux}} \) is the auxiliary question generated from the context \( \bm{P} \), and \( f_{\text{QA}} \) is the final QA model\footnote{we initialize our final QA model using the checkpoints of: \href{https://huggingface.co/csarron/bert-base-uncased-squad-v1}{https://huggingface.co/csarron/bert-base-uncased-squad-v1}, subsequently fine-tuned on Webis data.}.

\textbf{Task 2: Spoiler Generation with Title Text}
In this task, the document's title text is treated as a question. To increase the precision of the process, we integrate a BM25\footnote{https://pypi.org/project/rank-bm25/} retriever module to extract the top k paragraphs from the original content, thereby narrowing the context for spoiler generation. Details in Appendix \ref{sec:reduced}

\begin{equation}
\bm{S}_2 = 
\begin{cases}
f_{\text{QA}}(\bm{P}, \bm{Q}, \bm{A}), & \text{if } \bm{P}_{\text{top\_k}} = \bm{P} \\
f_{\text{QA}}(\bm{P}_{\text{top\_k}}, \bm{Q}, \bm{A}), & \text{if } \bm{P}_{\text{top\_k}} \subset \bm{P}
\end{cases}
\label{eq:sp_gen}
\end{equation}

Equation \ref{eq:sp_gen} indicates an alternative method for calculating \(\bm{S}_2\), where either the entire paragraph set \(\bm{P}\) or the reduced paragraph set \(\bm{P}_{\text{top\_k}}\) can be used for spoiler generation. The choice between these two depends on the specific use case and the desired balance between computational efficiency and accuracy.

We start by tokenizing both the original and generated questions ($\mathcal{Q}_{\text{orig}}$ and $\mathcal{Q}_{\text{gen}}$) alongside the contextual paragraphs ($\mathcal{P}$), utilizing a tokenizer $\tau$ that processes the text into token sequences while adhering to a maximum length of 384 tokens and ensuring proper padding and truncation. This results in sequences of token IDs and attention masks, as well as offset mappings ($\mathcal{T}_{\text{orig}}, \mathcal{O}_{\text{orig}}, \mathcal{T}_{\text{gen}}, \mathcal{O}_{\text{gen}}$). Concurrently, the answers ($\mathcal{A}$) are processed to ascertain their start and end positions within the tokenized context, translating character-level positions into token-level indices ($\text{start}_{\text{token}}, \text{end}_{\text{token}}$). Details in Appendix \ref{sec:qa}

The final spoiler \( \bm{S} \) is selected based on the combination of results from the main and auxiliary tasks, where the output spans with the lowest combined validation loss across both tasks are chosen as the final spoiler prediction

To optimize the parameters of our model, we use a loss function that combines the losses from both tasks. The total loss $\mathcal{L}$ is calculated as:
\begin{equation}
\mathcal{L} = \mathcal{L}_2 + \alpha \cdot \mathcal{L}_1
\end{equation}
where \( \mathcal{L}_1 \) and \( \mathcal{L}_2 \) are the cross-entropy (same as Eq. \ref{eq:cross_entropy_loss}) losses for Task 1 and Task 2, respectively, and \( \alpha \) is a hyperparameter. It is experimentally determined from the validation set. In our case,  \( \alpha \) value of $0.5$ yielded optimal results. Implementation details in Appendix \ref{sec:tr}

\section{Results and Analysis}

\subsection{Spoiler Classification Task}

We experimented with spoiler classification within a multi-class setting and rigorously tested many neural models, including DistillBERT\cite{Sanh2019DistilBERTAD}, RoBERTa\cite{zhuang-etal-2021-robustly}, and Longformer\cite{Beltagy2020LongformerTL} reported in Table \ref{table:sp_cls}. We found our RoBERTa-Large model surpassed the state-of-the-art result of accuracy 73.63\% \cite{hagen-etal-2022-clickbait}, albeit by a small margin and it has been statistically analyzed (Appendix \ref{sec:ttest}).

\begin{table}[]
\centering
\scriptsize
\begin{tabular}{@{}lrrrr@{}}
\toprule
\textbf{Model Name} &
  \multicolumn{1}{l}{\textbf{Eval Acc}} &
  \multicolumn{1}{l}{\textbf{Test Acc}} &
  \multicolumn{1}{l}{\textbf{Eval F1}} &
  \multicolumn{1}{l}{\textbf{Test F1}} \\ \midrule
DistilBERT    & 67.8           & 67.7        & 67.7           & 66.2           \\
Longformer    & 68.75          & 68.43       & 67.56          & 66.46          \\
RoBERTa       & 71.8           & 71.46       & 70.3           & 70.26          \\
RoBERTa-Large & \textbf{73.56} & \textbf{75} & \textbf{72.59} & \textbf{73.74} \\ 
\bottomrule
\end{tabular}
\caption{Comparison of model performances for Spoiler Classification Task}
\label{table:sp_cls}
\end{table}

\subsection{Spoiler Generation Task}

\subsubsection{MTL vs. Non-MTL}

In this section, we dive into the intricacies of spoiler generation and the implications of the methodologies employed in the study. The quality of generated spoilers is evaluated based on several metrics(Appendix~\ref{sec:evalm}).

From table \ref{table:phz},\ref{table:pssg} we observe Roberta-Large achieved the highest BLEU-4 score for phrase spoilers among BERT-based models, while LongT5 significantly outperformed others in passage spoiler generation despite a higher computation cost (Details in Appendix \ref{sec:longt5}).

\begin{table}[]
\centering
\scriptsize
\begin{tabular}{@{}llll@{}}
\toprule
\textbf{Model}         & \textbf{BLEU-4} & \textbf{METEOR} & \textbf{BERT Sc.} \\ \midrule
BERT-b-u (v)           & 56.96           & 47.44           & 76                \\
BERT-b-u (t)           & 52.79           & 59.16           & 84                \\
BERT-b-u (MTL) (v)     & 47.69           & 40.12           & 67                \\
\midrule
RoBERTa-b (v)          & 64.11           & 53.92           & 79                \\
RoBERTa-b (t)          & 61.26           & 53.82           & 78                \\
\midrule
Our RoBERTa-l (v) & 73.36  & 61.38  & 84       \\ 
Hagen et al. RoBERTa-l (v)(n=97)* & 79.47  & 78.61  & 84.04       \\ 
\textbf{Our RoBERTa-l (t)} & \textbf{68.35}  & 60.79  & \textbf{83} \\
Hagen et al. RoBERTa-l (t) & 65.70  & 66.15  & 74.81 \\ 
\bottomrule
\end{tabular}
\caption{Phrase Spoiler Generation Results——Here the letter b stands for the base, l stands for large, and c/u denotes cased or uncased, (v) means validation samples, (t) means test samples, Total phrase spoiler in validation samples were 335, test samples were 423. * represents Hagen et al. used a subset of 97 Phrase spoilers out of 335 to report the scores.}
\label{table:phz}
\end{table}

\begin{table}[]
\centering
\scriptsize
\begin{tabular}{@{}llll@{}}
\toprule
\textbf{Model}                 & \textbf{BLEU-4} & \textbf{METEOR}      & \textbf{BERT Score} \\ \midrule
BERT-b-u (v)                   & 21.11           & 22.72                & 52.68               \\
BERT-b-u (t)                   & 17.09*          & 23.65*(335)          & 53.62(335)          \\
BERT-b-u (MTL) (v)             & 14.28           & 13.21                & 44                  \\
BERT-b-u (MTL) (t)             & \textbf{17.78*}          & 22.56*(335)          & 53*(335)            \\
\midrule
RoBERTa-b (v)                  & 26.73           & 27.97                & 58                  \\
RoBERTa-b (t)                  & 21.81*          & 29.72*(335)          & 58*(335)            \\
\textbf{RoBERTa-b (MTL) (v)}   & \textbf{26.86}  & \textbf{28.14}       & 54                  \\
\textbf{RoBERTa-b (MTL) (t)}   & \textbf{22.59*} & \textbf{32.67*(335)} & \textbf{60(335)}    \\
\midrule
RoBERTa-l (v)                  & 30.67           & 32.78                & 58                  \\
RoBERTa-l (t)                  & 26.52           & 36.66(335)           & 63(335)             \\
RoBERTa-l (MTL) (v)            & 29.71           & 30.99                & 56                  \\
RoBERTa-l (MTL) (t)            & 26.65           & 36.42(335)           & 63(335)             \\
\midrule
Hagen et al. (Best)            & 31.44           & 46.06                & 51.06               \\
\textbf{LongT5 (v) (our best)} & \textbf{88.72}  & \textbf{90.29}       & \textbf{97.98}      \\
\textbf{LongT5 (t) (our best)} & \textbf{90.10}  & \textbf{90.81}       & \textbf{98.17}      \\ \bottomrule
\end{tabular}
\caption{Passage Spoiler Generation Results——Here the letter b stands for the base, l stands for large, and c/u denotes cased or uncased, (v) means validation samples, (t) means test samples, * means the setting includes reduction of context. Total passage spoilers in validation samples were 322, and test samples were 403, and Numbers in the brackets indicate a subset of size n (random sample)}
\label{table:pssg}
\end{table}

\subsubsection{Case Study: Does context reduction help in MTL?}
Our examination of 335 passage spoilers from a test dataset showed no notable improvements initially, prompting an investigation into how context size affects Multitask Learning (MTL) and non-MTL setups, as detailed in Table \ref{table:pssg}. Utilizing the retriever component revealed a distinct advantage for MTL setups, especially with the RoBERTa base model, pointing towards intriguing future research possibilities on context reduction and retriever models.

\subsubsection{Multi Spoilers}
We used fine-tuned LongT5 to generate multi-spoilers and found exceptionally good results in spoiler generation quality, reported in Table \ref{table:multi}.

\begin{table}[]
\centering
\scriptsize
\begin{tabular}{@{}llll@{}}
\toprule
\textbf{Model} & \textbf{BLEU-4} & \textbf{METEOR} & \textbf{BERT Score} \\ \midrule
Hagen et al.            & -           & -                & -               \\
LongT5         & 81.55           & 85.39           & 96.67               \\ \bottomrule             
\end{tabular}
\caption{Multi Spoiler Generation Results}
\label{table:multi}
\end{table}

\section{Conclusion and Future Work}
In this work, we have notably advanced the field of clickbait spoiling, particularly excelling in scenarios involving reduced context. Our research lays a solid groundwork for finding the optimal parameters for context reduction and innovative context generation techniques, ensuring better performance in spoiler generation tasks. Furthermore, the study underscores the superiority of LongT5-based models over their BERT-based counterparts in generating extended spoilers. The sensitivity of our MTL model's performance to the alpha hyperparameter, initially a limitation, opens a direction for developing adaptive strategies to fine-tune the balance between primary and auxiliary tasks, optimizing the model's efficiency. The future also holds the potential for rigorous statistical analysis to strengthen the validity of the performance gains attributed to Multitask Learning.


\bibliography{anthology,custom}
\bibliographystyle{acl_natbib}

\clearpage

\appendix

\section{Modified Preprocessing for Question Answering}
\label{sec:qa}
We have refined the conventional Question Answering (QA) mechanism, designed to generate spoilers from clickbait articles. This customization fundamentally resides in the preprocessing function, enabling separate QA for phrases and passages. We address the common issue of information loss due to the 512 tokens truncation limit of models like BERT by incorporating an overlapping sliding window strategy with a capacity of 384 tokens. The approach for determining the starting index of an answer depends on whether it's a phrase or a passage, with detailed strategies to handle different scenarios in case of passages. Our post-processing function retrieves the answer from the most confident window based on the logit score. Our innovative adaptation of the QA method has enabled us to effectively manage longer text sequences, thus enhancing the accuracy of spoilers generation and offering a robust solution to clickbait spoiling.

\section{MTL Training Details}
\label{sec:tr}

For the training procedure, PyTorch's DataLoader utility loads the shuffled training and validation datasets with a batch size of 8. We employ the AdamW optimizer, an extension of the Adam optimizer, with a learning rate 1e-5. A linear learning rate scheduler adjusts the learning rate based on the number of training steps computed over five epochs.
In each epoch, while in training mode, the model processes inputs for the original and auxiliary tasks separately, with each batch from the train-data loader. It computes the respective losses, combining them (with the auxiliary task's loss scaled down by a factor of 0.5) to backpropagate and update model parameters.
For the evaluation phase, the model, in evaluation mode, calculates the losses for both tasks using the eval-dataloader. It retains the start and end logits of each batch. After determining the average evaluation loss, the model checks if it is lower than the current minimum validation loss (initially set to a high value). If it is, the model state is saved, and the minimum validation loss is updated to the current average evaluation loss. This mechanism ensures that the best model, i.e., the one with the lowest validation loss, is retained.
Our MTL methodology concurrently processes two related tasks, improving learning efficacy and performance. This approach, valuable for tasks with substantial shared contexts, proves especially beneficial in our case of spoiler generation.

\section{Spoiler Generation in reduced context}
\label{sec:reduced}

In our research, we employed the LongT5 model, known for its effectiveness in language generation tasks, to perform zero-shot spoiler generation. This process involved reducing the context for each spoiler by retrieving the top k paragraphs which is similar to the generated spoiler. The top k (k=5 in our case) paragraphs with the highest similarity were selected to form the new context, effectively reducing the original context size while retaining the most pertinent information for spoiler generation.

The striking improvement in the quality of the generated spoilers, as observed within the Multitask Learning (MTL) framework, affirms the potential of our method in context reduction and spoiler generation. However, the experiment's outcomes also raise a pivotal question: What is the optimal number of paragraphs to include in the new context?

This query requires further exploration, and the answer will likely be a balancing act. On the one hand, including more paragraphs in the new context increases the probability of the actual spoiler being present within it but may pose a challenge for the Language model in accurately pinpointing the spoiler amidst the surplus of information. On the other hand, a shorter context is likely to aid the model in identifying the actual spoiler more efficiently, yet this depends on the reliability of the similarity metric to construct a context that sufficiently retains the spoiler and spoiler-like sentences.

Consequently, to optimally harness our approach's capabilities in tackling the clickbait spoiling challenge, future research should focus on exploring the optimal number of paragraphs to be included in the new context, considering both the efficacy of spoiler detection and the efficiency of the Language model.

\section{Generating extended spoilers}
\label{sec:longt5}
By adopting a fine-tuned Long-T5 model, we could handle both passage and multi-spoilers more effectively. Its ability to process longer sequence lengths led to superior results compared to the previous multitask learning approach, capturing more nuanced information within the texts. Training the Long-T5 model proved resource-intensive, requiring around 1.5 hours of training on an A100 40 GB GPU. Despite the higher computational demands and longer training times, the trade-off for significantly improved performance underscores the Long-T5 model's suitability for complex language tasks.

\section{Statistical analysis or significance tests - Roberta L Spoiler type classification}
\label{sec:ttest}
To ascertain the statistical significance of the performance improvement achieved by our RoBERTa-Large model in the spoiler classification task, we conducted a one-sample t-test. Our model's performance scores across five runs were compared against the previously reported state-of-the-art performance score of . The t-test yielded a t-statistic of 4.67 and a p-value of 0.009. Given the extremely low p-value, well below the conventional significance threshold of 0.01, we reject the null hypothesis that our model’s performance equals the state-of-the-art. This robustly indicates that the performance improvement is statistically significant and not a result of random variation. These results provide strong evidence that our model represents a meaningful advancement in the task of spoiler classification

\section{Evaluation Metrics}
\label{sec:evalm}

The evaluation matrices used for each are as follows
\begin{table}[h]
\centering
\begin{tabular}{|c|c|}
\hline
\textbf{Task Name} & \textbf{Metric} \\ 
\hline
Spoiler Classification & F1 \\
\hline
Spoiler Classification & Acc \\
\hline
Spoiler Generation & BLEU-4 \\
\hline
Spoiler Generation & METEOR \\
\hline
Spoiler Generation & BertScore \\
\hline
\end{tabular}
\caption{Performance metrics used for each task.}
\label{tab:my_label}
\end{table}

\end{document}